\documentclass[10pt,twocolumn]{article}

\usepackage[margin=0.72in]{geometry}
\usepackage{times}
\usepackage{microtype}
\usepackage{booktabs}
\usepackage{tabularx}
\usepackage{array}
\usepackage{makecell}
\usepackage{multirow}
\usepackage{amsmath}
\usepackage{amssymb}
\usepackage{enumitem}
\usepackage{xcolor}
\usepackage{hyperref}
\usepackage{url}
\usepackage{tikz}
\usepackage{graphicx}
\usepackage{pgfplots}
\usepackage{dblfloatfix}
\usepackage{float} 
\usepackage{placeins}
\usepackage{balance}
\usepackage{caption}
\usepackage{subcaption}

\usepackage{eso-pic}

\newcommand{\firstpagecorrespondence}{%
  \AddToShipoutPictureFG*{%
    \AtPageLowerLeft{%
      \raisebox{0.42in}{%
        \hspace{0.72in}%
        \parbox{0.86\paperwidth}{%
          \rule{\linewidth}{0.4pt}\\[-0.1em]
          {\footnotesize
          \textit{$^{*}$Corresponding authors:}
          \texttt{pratnaik@alumni.usc.edu};  
          \texttt{nareshapril8@gmail.com};  
          \texttt{yuew18@illinois.edu}
          }
        }%
      }%
    }%
  }%
}

\usetikzlibrary{arrows.meta, positioning, shapes.geometric, fit, calc, backgrounds}
\pgfplotsset{compat=1.18}

\hypersetup{
  colorlinks=true,
  linkcolor=blue,
  citecolor=blue,
  urlcolor=blue
}
\definecolor{accentblue}{RGB}{47,84,150}
\definecolor{accentgreen}{RGB}{53,122,83}
\definecolor{accentorange}{RGB}{191,112,28}
\definecolor{accentgray}{RGB}{92,98,110}
\definecolor{lightblue}{RGB}{239,244,251}
\definecolor{lightgreen}{RGB}{237,246,240}
\definecolor{lightorange}{RGB}{253,245,235}
\definecolor{lightgray}{RGB}{245,247,250}

\setlist[itemize]{leftmargin=*, topsep=2pt, itemsep=1pt}
\setlist[enumerate]{leftmargin=*, topsep=2pt, itemsep=1pt}

\newcommand{\inlineTableCaption}[2]{%
  \refstepcounter{table}\label{#1}%
  \noindent\textbf{Table~\thetable:} #2\par\vspace{0.25em}%
}

\title{\vspace{-1.0em}
Rethinking LLMOps for Fraud and AML: Building a Compliance-Grade LLM Serving Stack
\vspace{-0.5em}
}

\author{
Prathamesh Vasudeo Naik$^{*}$ \quad
Naresh Dintakurthi$^{*}$ \quad
Yue Wang$^{*}$\\
}

\date{}

\begin{document}
\firstpagecorrespondence
\twocolumn[
\maketitle
\vspace{-1.4em}
\begin{center}
{\large\bfseries Abstract\par}
\vspace{0.55em}
\begin{minipage}{0.78\textwidth}
\small
\setlength{\parindent}{0pt}
\setlength{\parskip}{0.55em}
Fraud detection and anti-money-laundering (AML) compliance are high-value domains for large language models (LLMs), but their serving requirements differ sharply from generic chat workloads. Compliance prompts are often prefix-heavy, schema-constrained, and evidence-rich: they include reusable policy instructions, risk taxonomies, transaction or document context, and short structured outputs such as JSON labels, risk factors, or analyst-facing summaries. These characteristics make prefix reuse, KV-cache efficiency, runtime tuning, multi-tenant model orchestration, and output-quality validation first-order systems concerns.

This paper introduces a workload-aware LLMOps stack for fraud and AML workloads, centered on self-hosted open-weight models such as Meta Llama and Alibaba Qwen. The stack combines vLLM-style runtime tuning, PagedAttention, Automatic Prefix Caching, multi-adapter serving, adapter- and prompt-length-aware batching, sleep/wake lifecycle management, speculative decoding, and optional prefill/decode disaggregation. To avoid exposing institution-specific data, our reproducibility track converts public synthetic AML datasets, including IBM AML and SAML-D, into prefix-heavy compliance prompts with reusable policy text, transaction evidence, typology definitions, and schema-constrained outputs.

We also incorporate an LLM-as-judge evaluation framework that uses deterministic compliance checks, reference metrics, expert-adjudicated calibration data where available, and multi-judge rubric scoring to select release-ready model, prompt, and adapter configurations. Across public-synthetic AML workloads derived from IBM AML and SAML-D, together with controlled serving benchmarks, workload-aware tuning improved throughput from 612--650 to 3,600 requests/hour, reduced P99 latency from 31--38 seconds to 6.4--8.7 seconds, and increased GPU utilization from 12\% to 78\%. In a multi-adapter compliance endpoint, adapter- and prompt-length-aware batching improved effective throughput by 3.90$\times$ over PEFT-style adapter loading. These results should be interpreted as serving-stack and workload-shape evidence on public synthetic data, not as disclosure of any proprietary financial-institution system or private investigation process.

\vspace{0.25em}
\end{minipage}
\end{center}
\vspace{1.0em}
]

\section{Introduction}

Large language models are increasingly used in fraud detection, risk review, and AML compliance to extract risk factors, classify predicate offenses, summarize case evidence, generate analyst-ready narratives, and support back-office investigation workflows. These applications are operationally attractive because much of the work involves reading semi-structured evidence, applying policy language, and producing structured or narrative outputs. However, deploying LLMs in regulated financial systems is not simply a matter of choosing the strongest model. Data locality, cost predictability, latency service-level agreements (SLAs), output validity, auditability, and operational control are equally important.

A recurring pattern in fraud and compliance workloads is that requests are \emph{prefix-dominated}. Each prompt may include a large shared instruction block, policy language, schema definitions, risk-factor taxonomies, or compliance-specific formatting rules, while only the case-specific evidence changes between requests. The output is often short structured JSON or a compact list of labels. This pattern differs from open-ended chat, where each conversation may have diverse history and long outputs. It also differs from classic offline summarization, where latency is less critical and shared prefixes may be less dominant.

This distinction changes the serving optimization order. Default LLM serving configurations often leave major performance on the table because they are not tuned to the shape of the workload. In our case studies, the largest gains came from practical systems choices: increasing active sequence concurrency without triggering preemption, sizing batched-token capacity to avoid split prefills, using paged KV memory to reduce fragmentation, enabling automatic prefix caching, restricting generation with stop tokens and realistic maximum-token limits, grouping requests by adapter and prompt length, and avoiding cold model loads through sleep/wake lifecycle control. Advanced techniques such as speculative decoding and prefill/decode disaggregation became useful after simpler workload-specific bottlenecks were understood.

This paper makes four contributions. First, we characterize fraud and compliance inference as a distinct workload class with long shared prefixes, structured short outputs, and frequent multi-tenant model usage. Second, we present a reusable serving-stack design that maps concrete optimization techniques to specific AML and fraud workloads. Third, we define a public reproducibility track by converting public synthetic AML datasets into prefix-heavy LLM serving workloads, avoiding the disclosure of institution-specific transaction or investigation data. The reproducible experiments use IBM AML as the primary public synthetic benchmark and SAML-D as a secondary academic benchmark; no proprietary customer records, account identifiers, employer-specific case files, internal investigation notes, or confidential transaction logs are used or disclosed. Fourth, we describe an LLM-as-judge quality gate for preproduction evaluation, where public golden references, expert calibration where available, and multi-judge scoring determine whether a serving configuration is eligible for rollout.

\section{Related Work}

LLM serving performance increasingly depends on systems design rather than model architecture alone. Recent surveys organize inference-serving advances around batching, scheduling, KV-cache management, memory hierarchy, parallelism, and distributed serving~\cite{li2024survey}. PagedAttention and vLLM introduced a virtual-memory-inspired KV-cache layout that reduces fragmentation and improves high-throughput serving for long and variable-length prompts~\cite{kwon2023pagedattention}. These techniques are directly relevant to fraud and compliance workloads, where prompts often contain long policy text, evidence summaries, schemas, and reusable task instructions.

Prefix reuse is especially important for repeated-prompt workloads. Prompt Cache and CachedAttention reuse attention states for modular or recurring prompt segments~\cite{gim2023promptcache,gao2024cachedattention}. vLLM's Automatic Prefix Caching reuses KV-cache blocks for matching prefixes, while LMCache extends reuse across GPU, CPU, disk, and remote tiers~\cite{vllmapc,lmcache,huang2025lmcache}. Related systems also disaggregate prefill and decode: Splitwise and DistServe show that separating compute-heavy prompt processing from memory-heavy token generation can improve goodput and reduce head-of-line blocking~\cite{patel2023splitwise,zhong2024distserve}.

Structured generation further motivates workload-aware serving. SGLang introduces RadixAttention for KV reuse across structured language-model programs~\cite{zheng2024sglang}. Speculative decoding accelerates generation by using draft tokens verified by the target model~\cite{leviathan2023speculative}, with EAGLE-3 improving the approach through direct token prediction and multi-layer feature fusion~\cite{li2025eagle3}. AIConfigurator highlights the growing complexity of manually tuning serving parameters and proposes automated configuration search across runtime knobs such as KV budget, CUDA graphs, and token capacity~\cite{aiconfigurator2026}.

LLM-as-a-judge methods are increasingly used to evaluate open-ended outputs, but they require safeguards. MT-Bench, Chatbot Arena, and G-Eval show that strong judges can align with human preferences in some settings~\cite{zheng2023judge,liu2023geval}. However, prior work also documents position bias, verbosity bias, and self-preference, motivating answer-order swaps, blinded comparisons, and heterogeneous judge panels~\cite{wang2024faireval,panickssery2024selfpref,verga2024poll}. These limitations are particularly important in fraud and AML, where unsupported but fluent narratives can create compliance risk.

Public AML datasets enable reproducible evaluation without exposing confidential financial-institution data. IBM Transactions for Anti Money Laundering provides a public synthetic financial-transaction benchmark with laundering labels~\cite{altman2023aml,ibmamlrepo}. SAML-D provides synthetic transaction-monitoring records with suspicious labels and typology metadata, but its CC BY-NC-SA 4.0 license makes it more appropriate as a secondary academic benchmark~\cite{oztas2023samld,samldkaggle,ccbynsa}. SynthAML provides an alert-level synthetic benchmark for future narrative and alert-review extensions~\cite{jensen2023synthaml}. In this paper, IBM AML and SAML-D are used to construct prefix-heavy compliance prompts for serving evaluation.

Open-weight model families are central to regulated deployment because they allow self-hosting, adapter-based customization, runtime control, and cache-aware optimization. Llama 3 and Qwen2.5 provide strong open-weight foundations for this setting~\cite{llama3,qwen25}. For fraud and AML, their main advantage is not only model quality, but the ability to integrate model selection, serving optimization, governance, and quality validation within a controlled deployment stack.

\section{Fraud and Compliance Workload Characterization}

Fraud and AML workloads exhibit several recurring properties. First, prompts are frequently long because they contain document extracts, transaction summaries, dispute comments, login or geolocation evidence, risk taxonomies, and output schemas. Second, large portions of the prompt are reused across requests: the same policy instructions, JSON schema, decision rubric, and examples often appear in every call. Third, outputs are short but must be valid: a malformed JSON object, missing label, or repeated risk factor can be operationally unusable. Fourth, use cases are multi-tenant: predicate-offense classification, risk-factor extraction, translation, category mapping, and narrative generation may share a base model but require different adapters or prompts. Fifth, regulated environments often prefer self-hosted serving because external APIs may introduce data-governance concerns, shared-SLA risk, policy-blocking behavior, or unpredictable cost.

Table~\ref{tab:phenotype} summarizes the workload taxonomy and maps each phenotype to the most relevant serving optimizations.

\begin{table}[H]
\centering
\scriptsize
\setlength{\tabcolsep}{3pt}
\renewcommand{\arraystretch}{1.05}
\caption{Fraud/compliance workload phenotypes and preferred serving choices.}
\label{tab:phenotype}
\begin{tabularx}{\columnwidth}{>{\raggedright\arraybackslash}p{0.36\columnwidth}X}
\toprule
\textbf{Workload signal} & \textbf{Preferred serving choice} \\
\midrule
Long shared policy prefix & APC, PagedAttention, prefix-aware routing, LMCache. \\
Long input, short JSON output & Batched-token tuning, stop tokens, strict \texttt{max\_tokens}, greedy decode, CUDA graphs. \\
Multi-turn analyst context & SGLang RadixAttention, LMCache, cache-aware routing. \\
Many task variants on one base & Multi-adapter serving with adapter-aware and length-aware batching. \\
Sequential multi-model pipeline & vLLM sleep/wake lifecycle control. \\
Long prefills mixed with short decodes & Prefill/decode disaggregation. \\
\bottomrule
\end{tabularx}
\end{table}

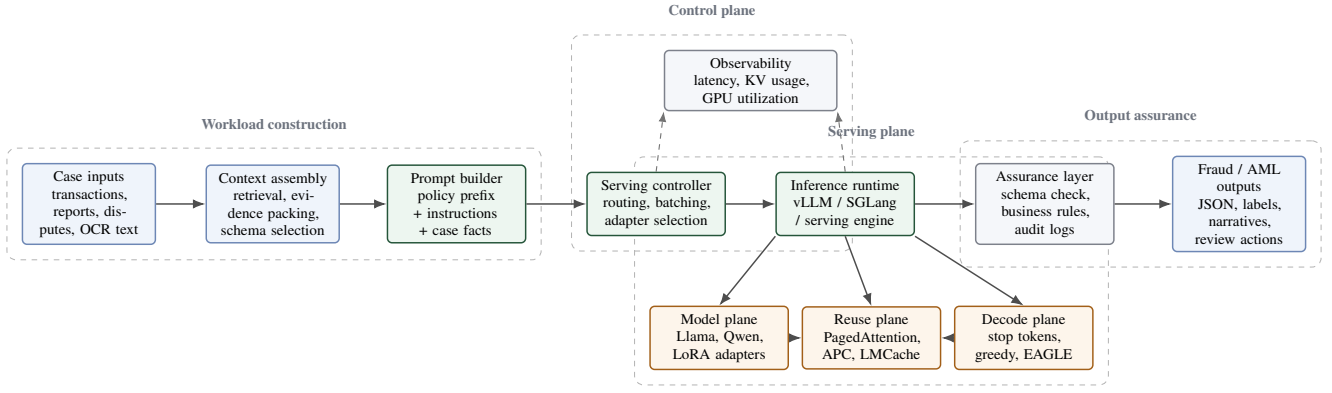
\begin{figure*}[!t]
\centering
\resizebox{0.97\textwidth}{!}{%
\begin{tikzpicture}[
  font=\footnotesize,
  every node/.style={align=center},
  data/.style={rectangle, rounded corners=2.5pt, draw=accentblue!70, fill=lightblue, thick, minimum height=0.82cm, text width=2.25cm, inner sep=4pt},
  proc/.style={rectangle, rounded corners=2.5pt, draw=accentgreen!70!black, fill=lightgreen, thick, minimum height=0.82cm, text width=2.35cm, inner sep=4pt},
  aux/.style={rectangle, rounded corners=2.5pt, draw=accentorange!85!black, fill=lightorange, thick, minimum height=0.78cm, text width=2.35cm, inner sep=4pt},
  guard/.style={rectangle, rounded corners=2.5pt, draw=accentgray!80, fill=lightgray, thick, minimum height=0.78cm, text width=2.35cm, inner sep=4pt},
  group/.style={rectangle, rounded corners=5pt, draw=black!30, dashed, inner sep=8pt},
  arr/.style={-{Latex[length=2.4mm]}, thick, draw=black!70},
  darr/.style={-{Latex[length=2.0mm]}, dashed, semithick, draw=black!55},
  lbl/.style={font=\bfseries\footnotesize, text=accentgray!90}
]

\node[data]  (src)     at (0,0)    {Case inputs\\transactions, reports, disputes, OCR text};
\node[data]  (prep)    at (3.5,0)  {Context assembly\\retrieval, evidence packing, schema selection};
\node[proc]  (prompt)  at (7.0,0)  {Prompt builder\\policy prefix + instructions + case facts};

\node[proc]  (router)  at (10.8,0) {Serving controller\\routing, batching, adapter selection};
\node[proc]  (runtime) at (14.4,0) {Inference runtime\\vLLM / SGLang / serving engine};

\node[guard] (post)    at (18.2,0) {Assurance layer\\schema check, business rules, audit logs};
\node[data]  (sink)    at (21.9,0) {Fraud / AML outputs\\JSON, labels, narratives, review actions};

\node[guard, text width=3.0cm] (obs) at (12.6,2.35) {Observability\\latency, KV usage, GPU utilization};

\node[aux] (model)  at (12.0,-2.55) {Model plane\\Llama, Qwen, LoRA adapters};
\node[aux] (kv)     at (14.9,-2.55) {Reuse plane\\PagedAttention, APC, LMCache};
\node[aux] (decode) at (17.8,-2.55) {Decode plane\\stop tokens, greedy, EAGLE};

\draw[arr] (src) -- (prep);
\draw[arr] (prep) -- (prompt);
\draw[arr] (prompt) -- (router);
\draw[arr] (router) -- (runtime);
\draw[arr] (runtime) -- (post);
\draw[arr] (post) -- (sink);

\draw[darr] (router.north) -- (obs.south west);
\draw[darr] (runtime.north) -- (obs.south east);

\draw[arr] (runtime.south west) -- (model.north);
\draw[arr] (runtime.south) -- (kv.north);
\draw[arr] (runtime.south east) -- (decode.north);

\draw[arr] (model.east) -- (kv.west);
\draw[arr] (decode.west) -- (kv.east);

\begin{scope}[on background layer]
  \node[group, fit=(src) (prep) (prompt)] (g1) {};
  \node[group, fit=(router) (obs)] (g2) {};
  \node[group, fit=(runtime) (model) (kv) (decode)] (g3) {};
  \node[group, fit=(post) (sink)] (g4) {};
\end{scope}

\node[lbl] at ($(g1.north)+(0,0.45)$) {Workload construction};
\node[lbl] at ($(g2.north)+(0,0.45)$) {Control plane};
\node[lbl] at ($(g3.north)+(0,0.45)$) {Serving plane};
\node[lbl] at ($(g4.north)+(0,0.45)$) {Output assurance};

\end{tikzpicture}%
}
\caption{Workload-aware serving architecture for prefix-dominated fraud and compliance inference. The stack separates workload construction, control, serving, and assurance planes so that prefix reuse, model tenancy, structured decoding, and output validation can be optimized jointly.}
\label{fig:architecture}
\end{figure*}

Figure~\ref{fig:architecture} shows the serving architecture used in our case studies. The serving router can select a model family, adapter, runtime configuration, and caching policy based on the workload. The key design principle is that the serving stack should preserve reusable work: shared policy prefixes should not be recomputed, repeated context should be retained across turns or workers where possible, and adapters should be multiplexed without repeatedly loading model weights.

\section{Benchmarking Methodology}

The benchmark harness is designed around \emph{schema-valid goodput}, not raw tokens per second alone. A response counts as successful only if it satisfies the target schema and passes task-specific post-checks, such as valid JSON, non-empty risk labels, and no duplicate risk-factor rows. For preproduction release decisions, we also evaluate semantic quality against expert-adjudicated calibration data using deterministic checks, reference metrics, and LLM-as-judge scores. This matters in regulated workflows because a malformed or unsupported output is operationally equivalent to failure.

For each optimization, we recommend preserving the empirical workload distribution: prompt-length histograms, output-length histograms, prefix-sharing ratios, and adapter mix should match production as closely as possible. We run both closed-loop concurrency sweeps and open-loop arrival tests, because settings that maximize throughput under closed-loop load can have poor P99 latency under bursty production traffic. Each configuration is warmed up before measurement, repeated at least three times, and reported with median and range.

Table~\ref{tab:protocol} lists the primary sweep space.

\begin{table}[H]
\centering
\scriptsize
\setlength{\tabcolsep}{3pt}
\renewcommand{\arraystretch}{1.05}
\caption{Benchmark protocol and configuration sweep space.}
\label{tab:protocol}
\begin{tabularx}{\columnwidth}{>{\raggedright\arraybackslash}p{0.31\columnwidth}X}
\toprule
\textbf{Component} & \textbf{Protocol} \\
\midrule
Warm-up / repeats & 50--100 requests or 5 min warm-up; 3 runs; report median and range. \\
Window & 10--15 min steady-state measurement per point. \\
Primary metrics & P50/P95/P99, TTFT, ITL, tokens/s, req/hour, GPU utilization, KV usage, OOM/preemption, schema-valid goodput. \\
Cost metric & GPU-hours per 1k successful requests. \\
Runtime sweeps & \texttt{max\_num\_seqs}, \texttt{max\_num\_batched\_tokens}, \texttt{gpu\_memory\_utilization}, \texttt{swap\_space}. \\
Decode/cache sweeps & Stop tokens, \texttt{max\_tokens}, greedy vs. sampling, CUDA graphs, APC, LMCache, speculative settings. \\
Multi-tenant sweeps & Adapter strategy/count, grouping policy, batch mixing, and prefill/decode resource split. \\
\bottomrule
\end{tabularx}
\end{table}

\paragraph{Public AML prompt construction.}
To reduce legal, privacy, and governance risk, the reproducibility track is built from public synthetic AML datasets rather than institution-specific transaction logs. IBM AML is used as the primary public benchmark because it is synthetic, publicly available, and includes laundering labels suitable for suspicious-flow and transaction-window tasks~\cite{altman2023aml,ibmamlrepo}. SAML-D is used as a secondary academic benchmark because it provides typology-rich synthetic transaction-monitoring records with suspicious labels and typology metadata~\cite{oztas2023samld,samldkaggle}. We do not use proprietary customer records, account identifiers, employer-specific case files, internal investigation notes, or confidential transaction data in the reproducible evaluation.

Each record or transaction window is rendered as a compliance prompt containing a reusable policy prefix, typology definitions, transaction evidence, and a required JSON schema. The public label becomes the golden reference for suspicious-pattern labeling, typology classification, or risk-factor extraction. Any expert-adjudicated calibration data is used only for local quality-gate design and is not required to reproduce the public serving benchmark.

\begin{table}[H]
\centering
\scriptsize
\setlength{\tabcolsep}{2.1pt}
\renewcommand{\arraystretch}{1.05}
\caption{Public AML datasets used for reproducible prompt construction. IBM AML is the primary public synthetic benchmark; SAML-D is used as a secondary academic benchmark due to its non-commercial license.}
\label{tab:publicdata}
\begin{tabularx}{\columnwidth}{>{\raggedright\arraybackslash}p{0.20\columnwidth}X>{\raggedright\arraybackslash}p{0.23\columnwidth}}
\toprule
\textbf{Dataset} & \textbf{Role in this paper} & \textbf{Data-use note} \\
\midrule
IBM AML~\cite{altman2023aml,ibmamlrepo} &
Primary transaction-window benchmark for suspicious-flow labeling, typology classification, and long-prefix JSON extraction. &
Synthetic data; CDLA-Sharing-1.0 data license~\cite{cdlasharing}. \\
SAML-D~\cite{oztas2023samld,samldkaggle} &
Secondary typology-rich benchmark for risk-factor extraction, suspicious-pattern labeling, and adapter-specific task batching. &
Synthetic data; CC BY-NC-SA 4.0~\cite{ccbynsa}. \\
SynthAML~\cite{jensen2023synthaml} &
Alert-level public benchmark referenced as a compatible extension for future narrative and alert-review prompts. &
Synthetic alert-level benchmark. \\
\bottomrule
\end{tabularx}
\end{table}

The experiments below mix public-synthetic AML prompt workloads for reproducibility with controlled serving microbenchmarks for runtime features such as speculative decoding, multi-adapter batching, and sleep/wake lifecycle control. The same harness separates serving measurements from quality-gating measurements so that faster configurations are not promoted unless they remain compliant and judge/human aligned.

\subsection{LLM-as-Judge Quality Gate}
\label{sec:judge}

Serving efficiency is necessary but not sufficient for fraud and AML deployment. A faster model, prompt, or adapter should be promoted only if it remains aligned with investigator expectations and compliance policy. We therefore use an LLM-as-judge framework as a preproduction quality gate. Prior AML agentic systems have used validation agents to improve SAR narrative quality and preserve investigator oversight~\cite{naik2025coinvestigator}; our framework generalizes this idea into a release gate for serving-stack changes, prompt revisions, model selection, and adapter updates. The framework is outside the online inference path: it evaluates candidate outputs before rollout, supports shadow-mode monitoring, and provides an auditable release decision, but it does not replace human investigators or final compliance review.

The workflow begins with golden-data construction. For the public reproducibility track, golden references come from public dataset labels such as laundering tags, suspicious flags, and typology fields. For local preproduction calibration, domain reviewers may additionally review sampled cases and record expected structured outputs, labels, narratives, or rubric scores. Such local calibration data is used only to validate the quality-gate methodology and is not included in the reproducible benchmark. Each evaluation record stores the task input, policy version, output schema, reference answer when available, required fields, and risk-slice metadata such as typology, language, synthetic evidence length, or long-context bucket. This design lets the same benchmark harness evaluate multiple candidate models, prompts, decoding configurations, and LoRA adapters without exposing proprietary investigation data.

The evaluator stack is deliberately multi-layered. First, deterministic checks enforce hard constraints such as valid JSON, required-field completeness, label whitelist compliance, evidence-citation presence, and prohibited-field absence. Second, reference-based metrics provide inexpensive signals when a golden answer exists: cosine similarity captures semantic proximity on normalized narratives, BLEU captures lexical precision for templated outputs, and ROUGE captures coverage for summary-like outputs \cite{papineni2002bleu,lin2004rouge,sbert2019}. Third, a panel of LLM judges scores the output against an investigator-written rubric. Judges produce structured scores, a short rationale, and a confidence or abstention field. For underspecified narrative tasks, pairwise comparison against the incumbent model is preferred over absolute scoring.

Judge selection is itself a preproduction experiment. Candidate judges are compared against adjudicated human labels using correlation with investigator scores, agreement on pass/fail or escalate decisions, and slice-level failure analysis. We also test robustness under answer-order swaps, blinded model identities, verbosity-controlled paraphrases, OCR-corrupted evidence, and rare predicate-offense categories. The production judge is selected only if it is stable, human-aligned, and non-inferior to the incumbent on high-risk slices. When judges disagree or confidence is low, the framework routes the item to human review and appends the adjudicated case to the next golden-data refresh.

Operationally, the release gate has six stages: investigator adjudication, deterministic schema and policy checks, reference-metric scoring, multi-judge rubric scoring, promotion or rollback decisioning, and shadow/canary monitoring after deployment. Each stage logs dataset, prompt, model, adapter, decoding configuration, judge version, and policy version so that evaluation decisions remain auditable.

This quality gate changes how serving results should be interpreted. The best deployment candidate is not simply the configuration with the lowest latency or highest throughput; it is the point on the serving-quality Pareto frontier that satisfies deterministic compliance checks and remains human-aligned under the judge framework. In regulated settings, this distinction is important because optimization can otherwise reward short, cheap, or highly formatted outputs that are not sufficiently faithful to case evidence.

\FloatBarrier
\section{Results}

All results in this section are reported on public-synthetic AML prompt workloads and controlled serving benchmarks. The public workloads are derived from IBM AML and SAML-D by rendering transaction records, labels, and typology fields into schema-constrained compliance prompts. The controlled serving benchmarks isolate runtime behaviors such as batching, prefix reuse, speculative decoding, multi-adapter execution, and sleep/wake lifecycle control. No proprietary customer, account, transaction, investigation, employer-specific, or internal compliance data is used in the reproducible evaluation.

\subsection{Public-Synthetic Workload Coverage}

Table~\ref{tab:publiccoverage} summarizes the public reproducibility slices used to stress the serving stack. The goal is to make prompt shapes, schema constraints, and label sources reproducible without exposing confidential records.

\begin{table}[H]
\centering
\scriptsize
\setlength{\tabcolsep}{2.2pt}
\renewcommand{\arraystretch}{1.04}
\caption{Public-synthetic AML prompt slices used in the reproducibility track. Token counts are measured after rendering the shared policy prefix, transaction evidence, and JSON schema.}
\label{tab:publiccoverage}
\begin{tabularx}{\columnwidth}{>{\raggedright\arraybackslash}p{0.22\columnwidth}>{\raggedright\arraybackslash}p{0.28\columnwidth}ccX}
\toprule
\textbf{Source} & \textbf{LLM task} & \makecell{\textbf{Avg.}\textbf{input}} & \makecell{\textbf{Avg.}\textbf{output}} & \textbf{Reference signal} \\
\midrule
IBM AML & Suspicious-flow labeling and transaction-window extraction & 2.1k & 72 & Synthetic laundering label and transaction pattern. \\
SAML-D & Typology classification and risk-factor extraction & 1.6k & 68 & Suspicious flag, payment features, and typology field. \\
Mixed public slice & Multi-task adapter benchmark across classification, extraction, and escalation tags & 1.2k--2.8k & 64--96 & Dataset label plus schema-valid JSON checks. \\
\bottomrule
\end{tabularx}
\end{table}

\subsection{Runtime Tuning for Long-Context Extraction}

The strongest serving result came from a public-synthetic long-context extraction workload in which serialized transaction evidence, reusable policy text, and schema-constrained LLM inference shared a constrained A100-class serving environment. Baseline telemetry showed high P99 latency even though GPU utilization remained low. Profiling identified several interacting causes: too few active sequences, split prefills caused by an overly small batched-token budget for roughly 2.5k-token inputs, conservative KV allocation, excessive swap reservation, and avoidable decode overhead.

The most important sweep was \texttt{max\_num\_seqs}. In our measured run, \texttt{max\_num\_seqs=16} underutilized the device, \texttt{max\_num\_seqs=32} caused frequent preemption and OOM behavior, and \texttt{max\_num\_seqs=24} was the best operating point. Increasing \texttt{max\_num\_batched\_tokens} above the workload's common prefill length reduced split prefills. Reducing \texttt{swap\_space} from 16 GiB to 4 GiB recovered host memory without harming stability. Setting \texttt{enforce\_eager=False} enabled CUDA graphs and reduced overhead.

\begin{table}[H]
\centering
\footnotesize
\setlength{\tabcolsep}{3pt}
\caption{Case A: public-synthetic long-context extraction workload before and after workload-aware tuning. Values are controlled serving measurements.}
\label{tab:casea}
\begin{tabularx}{\columnwidth}{Xrr}
\toprule
\textbf{Metric} & \textbf{Baseline} & \textbf{Optimized} \\
\midrule
Throughput & 612--650 req/hr & 3,600 req/hr \\
P99 latency & 31--38 s & 6.4--8.7 s \\
Average latency & 18.2 s & $<$9 s \\
GPU utilization & 12\% & 78\% \\
Normalized cost & 3.08--3.27 & 0.56 \\
Capacity plan & $\sim$10 GPUs & $\sim$3--4 GPUs \\
\bottomrule
\end{tabularx}
\vspace{0.2em}
\caption*{\footnotesize Normalized cost is A100-hours per 1k successful requests.}
\end{table}

Table~\ref{tab:casea} shows that the optimized configuration increased throughput by roughly 5.5--5.9$\times$, reduced P99 latency to single digits, and raised GPU utilization from 12\% to 78\%. Using the same serving footprint, GPU-hours per 1k successful requests fell by approximately 83\%.
\FloatBarrier

\subsection{Decode Control and Speculative Decoding}

Many compliance outputs are short JSON objects, so excessive output budgets are wasteful. In one extraction workload, typical output length was approximately 50--75 tokens while \texttt{max\_tokens} was set to 200. Adding a task-specific stop marker saved roughly 125 tokens per request. Because the task already used temperature 0, disabling unnecessary sampling overhead reduced per-token processing without changing semantics.

Speculative decoding became useful after basic decode cleanup. In a guided-JSON H100 benchmark, an EAGLE-style speculative configuration on 1$\times$H100 matched or slightly exceeded a 2$\times$H100 NIM baseline at the measured operating point, while reducing normalized GPU-hours per 1k requests by roughly half.

\begin{table}[H]
\centering
\footnotesize
\setlength{\tabcolsep}{3pt}
\caption{Case B: guided-JSON public-synthetic compliance benchmark with speculative decoding. Values are controlled serving measurements.}
\label{tab:caseb}
\begin{tabularx}{\columnwidth}{Xrrr}
\toprule
\textbf{Configuration} & \textbf{GPUs} & \textbf{Throughput} & \textbf{Latency} \\
\midrule
NIM baseline & 2$\times$H100 & 3.01 req/s & 2397 ms \\
vLLM + EAGLE-style & 1$\times$H100 & 3.19 req/s & 2292 ms \\
vLLM + EAGLE-style & 2$\times$H100 & 3.60 req/s & 2067 ms \\
\bottomrule
\end{tabularx}
\vspace{0.2em}
\caption*{\footnotesize Normalized cost was 0.185, 0.087, and 0.154 H100-hours per 1k requests, respectively.}
\end{table}

The 1$\times$H100 speculative configuration reduced normalized cost by about 53\% relative to the 2$\times$H100 baseline. Controlled tuning also showed that \texttt{draft\_tensor\_parallel\_size=1} was preferable for most latency-sensitive runs; \texttt{=2} was useful only under sustained high concurrency and introduced higher variance.
\FloatBarrier

\subsection{Prefix Reuse and KV Management}

Prefix reuse is the most domain-specific optimization in our study. In the public-synthetic long-prefix prompt track, moving from large contiguous KV allocation to paged KV layout reduced effective memory waste from roughly 60\% to under 4\%. Combined with APC for repeated policy and schema prefixes, this enabled a broader throughput improvement path from roughly 250k daily prompts toward multi-million daily prompt capacity without linear hardware growth.

APC is most effective when the workload repeatedly uses the same prefix. In fraud and compliance, this condition is common: the system prompt, policy text, JSON schema, few-shot examples, and risk taxonomy may remain identical across thousands of requests. LMCache is the natural escalation path when reuse must persist across engine replicas, workers, or cold/warm memory tiers. We therefore treat LMCache as an advanced extension of APC rather than a replacement.
\FloatBarrier

\subsection{Model Choice: Llama and Qwen for AML Tasks}

The quality pilots show that model choice should be workload-specific. Llama-3.1-8B was strong among open-weight models for predicate-offense extraction and was repeatedly preferred for stable JSON and finer-grained risk-factor decomposition. Qwen2.5 models offered an attractive cost and latency ladder, with useful precision in budget-sensitive extraction tasks even when recall trailed the strongest closed models in small pilots.

\begin{table}[H]
\centering
\scriptsize
\setlength{\tabcolsep}{3pt}
\renewcommand{\arraystretch}{1.05}
\caption{Public-synthetic AML model-selection observations from pilot task slices.}
\label{tab:modelquality}
\begin{tabularx}{\columnwidth}{>{\raggedright\arraybackslash}p{0.34\columnwidth}X}
\toprule
\textbf{Task slice} & \textbf{Pilot observation and serving implication} \\
\midrule
Predicate-offense extraction & Llama-3.1-8B produced stable labels/JSON (about 85\% recall and precision on a four-sample pilot), making it a strong self-hosted default. \\
Budget-tier extraction & Qwen2.5 7B/14B/32B produced conservative labels (about 50\% recall / 85\% precision), useful for low-cost precision-oriented stages. \\
Risk-factor extraction & Llama decomposed broader factors into granular rows and maintained more stable JSON, while API baselines sometimes had higher coverage. \\
Risk-factor naming / clustering & Larger API models were stronger on some small recall-critical slices; use hybrid review or larger models for these tasks. \\
External API comparison & API models can be useful quality probes, but policy refusals, resource limits, and JSON parse failures motivate regulated self-hosted serving paths. \\
\bottomrule
\end{tabularx}
\end{table}

The key conclusion is not that Llama or Qwen universally beat hosted API models. Rather, they enable a controllable self-hosted stack where caching, adapters, scheduling, and governance can be jointly optimized. This matters in AML because prompts may legitimately contain suspicious or policy-sensitive content, and because regulated deployments require predictable latency and output formatting.

\subsection{Multi-Adapter and Length-Aware Batching}

Compliance platforms commonly serve several related tasks over the same base model: category classification, policy classification, translation, risk-factor extraction, buyer-intent labeling, and reasoning tags. LoRA adapters are attractive because a single Llama- or Qwen-class base model can support many task-specific behaviors without loading a separate full model for every task. The serving challenge is that a production request may itself contain a mini-batch of prompts destined for different adapters. A naive implementation processes those prompts sequentially, or sends each prompt as an independent request to the endpoint, losing both batching efficiency and adapter locality.

Our serving strategy uses two levels of grouping. First, the endpoint partitions the incoming request by adapter identifier, so all prompts that require the same LoRA weights are executed together. Second, within each adapter group, prompts are bucketed by approximate input length. This second step matters because fraud and compliance workloads mix short classification prompts, medium evidence summaries, and long OCR/report extraction prompts. If a 400-token classification prompt and a 3,000-token extraction prompt are placed in the same prefill batch, the shorter request inherits padding and scheduling inefficiency. Adapter-first grouping preserves model specialization; length-aware grouping reduces prefill imbalance.

The dispatch procedure is simple enough to run inside the serving controller: parse each item into \texttt{(adapter\_id, prompt\_len, original\_index)}, group by \texttt{adapter\_id}, sub-bucket by length ranges such as short, medium, and long, execute each bucket as a batch call, and finally restore the original order before returning the response. This is especially useful for back-office batch APIs, where a single upstream compliance workflow may submit dozens of heterogeneous prompts but still expects a response array in the original order.

\begin{table}[H]
\centering
\scriptsize
\setlength{\tabcolsep}{2.4pt}
\renewcommand{\arraystretch}{1.05}
\caption{Case C1: mixed 12-prompt compliance request across four LoRA adapters. Throughput is $12/\mathrm{latency}$.}
\label{tab:casec_adapter}
\begin{tabularx}{\columnwidth}{>{\raggedright\arraybackslash}p{0.34\columnwidth}rrr}
\toprule
\textbf{Serving strategy} & \textbf{Lat.} & \textbf{Tput} & \textbf{Gain} \\
\midrule
PEFT adapter loading & 8.916 s & 1.35 req/s & 1.00$\times$ \\
LookAhead adapter loading & 4.420 s & 2.71 req/s & 2.01$\times$ \\
Merge-and-save + PEFT & 6.188 s & 1.94 req/s & 1.44$\times$ \\
vLLM multi-adapter batch & 2.840 s & 4.23 req/s & 3.13$\times$ \\
\textbf{vLLM multi-adapter + length buckets} & \textbf{2.280 s} & \textbf{5.26 req/s} & \textbf{3.90$\times$} \\
\bottomrule
\end{tabularx}
\vspace{0.15em}
\caption*{\scriptsize Request split: A1=4 policy/category prompts, A2=3 risk-factor prompts, A3=3 buyer-intent prompts, A4=2 reasoning prompts.}
\end{table}

Table~\ref{tab:casec_adapter} reports a single controlled benchmark for a mixed compliance request. A single upstream request may contain an arbitrary number of prompts per adapter; in this benchmark the request contained four adapters with a 4/3/3/2 prompt split. The serving controller first groups items by \texttt{adapter\_id}, then buckets each adapter group by input length. Native vLLM multi-adapter batching reduced batch latency from 8.916 seconds to 2.840 seconds by avoiding repeated adapter-loading and preserving base-model residency. Adding prompt-length buckets further reduced latency to 2.280 seconds and increased effective throughput from 1.35 to 5.26 requests/s, a 3.90$\times$ throughput gain over the PEFT-loading baseline. The operational lesson is that multi-adapter inference should not be treated only as a memory-saving feature; it is also a batching and routing problem where adapter locality and prompt-length locality jointly determine serving efficiency.

\subsection{Sleep/Wake Lifecycle for Sequential Model Pipelines}

Sleep/wake lifecycle control addresses a different production pattern. Some fraud and compliance workflows are sequential rather than parallel: an extraction model prepares evidence, a classifier model maps it to a typology or predicate offense, and a larger reviewer model may be invoked only for escalations. Keeping all models resident at once wastes VRAM, but cold-loading each model at every stage adds tens of seconds of availability delay. Sleep mode provides a middle ground: inactive models release enough GPU memory for the next stage while preserving a much faster wake path than a full cold start.

\begin{table}[H]
\centering
\footnotesize
\setlength{\tabcolsep}{3pt}
\renewcommand{\arraystretch}{1.06}
\caption{Case C2: measured cold-start versus sleep/wake availability for sequential model stages.}
\label{tab:casec_sleep}
\begin{tabularx}{\columnwidth}{Xrrr}
\toprule
\textbf{Model stage} & \textbf{Cold} & \textbf{Wake} & \textbf{Reduction} \\
\midrule
Llama-8B extractor & 36.39 s & 0.98 s & 37.1$\times$ \\
Qwen-14B classifier & 79.09 s & 2.65 s & 29.8$\times$ \\
Qwen-32B reviewer & 125.89 s & 4.15 s & 30.3$\times$ \\
\bottomrule
\end{tabularx}
\end{table}

Table~\ref{tab:casec_sleep} shows that sleep/wake reduced model availability time by roughly 30--37$\times$ compared with cold starts. This result is separate from multi-adapter batching: adapters are best when many tasks share one base model, while sleep/wake is best when different full models must run in sequence but not simultaneously. In our serving playbook, the two methods therefore target different bottlenecks: adapter batching reduces intra-endpoint fragmentation, and sleep/wake reduces inter-stage cold-start churn.

\FloatBarrier
\section{Operational Playbook}
Table~\ref{tab:playbook} summarizes the operational playbook derived from the case studies. The table intentionally starts from observable telemetry, because production engineers rarely know the root cause at the beginning of an incident or scaling effort.

\begin{center}
\begingroup
\scriptsize
\setlength{\tabcolsep}{3pt}
\renewcommand{\arraystretch}{1.05}
\inlineTableCaption{tab:playbook}{Bottleneck-to-intervention decision table for fraud and compliance serving.}
\begin{tabularx}{\columnwidth}{>{\raggedright\arraybackslash}p{0.38\columnwidth}X}
\toprule
\textbf{Telemetry signal} & \textbf{Diagnosis and intervention} \\
\midrule
GPU utilization $<$20\%, high latency & Too few active sequences or split prefills; tune \texttt{max\_num\_seqs} and \texttt{max\_num\_batched\_tokens}; then try APC/CUDA graphs. \\
Frequent preemption or OOM & Over-aggressive concurrency or KV budget; reduce sequence count and tune \texttt{gpu\_memory\_utilization}/\texttt{swap\_space}. \\
Long common prefix & Redundant prefill; enable APC, then LMCache or prefix-aware routing if reuse crosses workers. \\
Short JSON but high decode time & Avoidable generation overhead; use stop tokens, lower \texttt{max\_tokens}, greedy decode, then EAGLE/speculation. \\
Many adapters on one base & Adapter/length fragmentation; use multi-adapter serving with adapter-aware and length-aware batching. \\
Sequential model stages & Cold-start and VRAM churn; use sleep/wake lifecycle with trusted backend control. \\
Long prefills block short decodes & Head-of-line blocking; evaluate prefill/decode disaggregation and cache-transfer optimization. \\
Judge-human disagreement & Quality risk; route to investigator review, refresh golden data, recalibrate the judge panel, or block rollout. \\
\bottomrule
\end{tabularx}
\endgroup
\end{center}

\FloatBarrier
\section{Limitations and Future Work}

This study has four limitations. First, the public reproducibility track relies on synthetic AML datasets rather than confidential institution-specific data, so it cannot capture every operational nuance of real investigations. The results should therefore be interpreted as serving-stack and workload-shape evidence, not as disclosure of any private financial-institution performance, customer behavior, investigation process, employer-specific workflow, or internal compliance system. IBM AML is used as the primary public benchmark because it is synthetic and released under a data-sharing license; SAML-D is used as a secondary academic benchmark because its CC BY-NC-SA license restricts commercial use~\cite{ccbynsa}. Second, the AML quality observations are small pilot slices intended for operational triage, not claims of state-of-the-art task quality. Third, LLM-as-judge evaluation is useful but imperfect: judges can exhibit position bias, verbosity bias, family preference, and drift under policy or task changes, so human adjudication and periodic recalibration remain necessary. Fourth, advanced extensions such as LMCache-backed cross-worker reuse and mature prefill/decode disaggregation are included as high-confidence next steps supported by prior work and early exploration, but not as the most stable production defaults in every setting.

Future work should expand the benchmark suite to larger labeled AML datasets, evaluate recall/precision trade-offs across more predicate-offense categories, and add automated configuration search to replace manual parameter sweeps. Another promising direction is joint routing across model family, adapter, cache locality, and quality risk: a request should be routed not only to the fastest worker, but also to the model and judge configuration most likely to pass the task-specific quality gate. Finally, scale-out experiments should quantify when prefill/decode disaggregation beats aggregated serving for mixed fraud traffic, and when active-learning loops should add judge-disagreement cases back into the golden dataset.

\section{Conclusion}

Fraud and compliance workloads are a distinct class of LLM serving problem. They combine long shared prefixes, structured short outputs, heterogeneous tenants, strict operational constraints, and high-stakes quality requirements. The main finding of this paper is that serving choices must follow the workload signature and remain inside a validated quality envelope. APC and PagedAttention are first-class tools for policy-grounded inference; runtime knobs such as \texttt{max\_num\_seqs}, \texttt{max\_num\_batched\_tokens}, \texttt{swap\_space}, and CUDA graphs should be tuned against real prompt distributions; speculative decoding is valuable once decode remains material after basic cleanup; multi-adapter and prompt-length-aware batching enables multi-skill compliance endpoints, while sleep/wake lifecycle control enables sequential multi-model pipelines; and prefill/decode disaggregation plus automated configuration search are natural next frontiers for scale-out regulated workloads.

In public-synthetic AML prompt workloads and controlled serving case studies, these choices delivered approximately 5.5--5.9$\times$ throughput gains, reduced P99 latency from tens of seconds to single digits, increased GPU utilization from 12\% to 78\%, and reduced projected capacity requirements from roughly 10 GPUs to roughly 3--4. These measurements are intended to characterize workload-aware serving behavior on public synthetic and controlled benchmarks, not to disclose proprietary enterprise performance. The broader implication is simple: for regulated inference, performance is a workload-design, quality-gating, and governance problem as much as a model or hardware problem.

\end{document}